\pdfoutput=1
\PassOptionsToPackage{table,dvipsnames}{xcolor}
\documentclass[11pt]{article}

\usepackage[final]{latex/acl}

\usepackage{times}
\usepackage{latexsym}
\usepackage[table]{xcolor}
\usepackage[dvipsnames]{xcolor}
\usepackage{makecell}

\usepackage[T1]{fontenc}

\usepackage[utf8]{inputenc}

\usepackage{microtype}

\usepackage{inconsolata}

\usepackage{inconsolata}
\usepackage{latexsym}
\usepackage{float}
\usepackage{mathrsfs}
\usepackage{amsthm,amsmath,amssymb}
\usepackage{multirow}
\usepackage[ruled]{algorithm2e}
\usepackage{threeparttable}
\usepackage{array}
\usepackage{booktabs}
\usepackage{bm}
\usepackage{graphicx}
\usepackage{amssymb}
\usepackage{wasysym}
\usepackage{bbm}

\definecolor{Mycolor1}{HTML}{FDC3D2}
\definecolor{Mycolor2}{HTML}{DCFDC4}
\title{Lower Layers Matter: Alleviating Hallucination via Multi-Layer Fusion and
Truthfulness Refocused Contrastive Decoding}

\author{
Dingwei Chen{$^{1,2}$}, Feiteng Fang{$^{4}$}, Shiwen Ni{$^{4*}$}, Feng Liang{$^{2}$}, Xiping Hu{$^{2}$},\\  \textbf{Ahmadreza Argha}{$^3$}, \textbf{Hamid Alinejad-Rokny}{$^3$}, \textbf{Min Yang}$^{4*}$, \textbf{Chengming Li}{$^{2}\thanks{~~Corresponding author.}$} \\
\small $^1$ Sun Yat-Sen University $^2$Shenzhen MSU-BIT University
$^3$UNSW, Sydney, NSW 2052, Australia \\
\small $^4$Shenzhen Key Laboratory for High Performance Data Mining, Shenzhen Institutes of Advanced Technology, Chinese Academy of Sciences 
\\ \small \texttt{chendw26@mail2.sysu.edu.cn, licm@smbu.edu.cn, min.yang@siat.ac.cn}
  }
  
\begin{document}
\maketitle
\begin{abstract}
Large Language Models (LLMs) have demonstrated exceptional performance across various natural language processing tasks. However, they occasionally generate inaccurate and counterfactual outputs, a phenomenon commonly referred to as ``hallucinations''. To tackle this issue, recent studies have explored contrastive decoding between the original model and an amateur model with induced hallucination, showing promising results. Nevertheless, this approach can disrupt  the original LLM's output distribution due to coarse contrast and simple subtraction operations, potentially leading to errors. In this paper, we introduce a novel contrastive decoding framework, termed \textbf{LOL} (\textbf{\underline{L}}\textbf{\underline{O}}wer \textbf{\underline{L}}ayer Matters). Unlike prior methods that focus solely on the final layer, our approach integrates contrastive information from lower layers to enable multi-layer fusion during contrastive decoding. Additionally, we incorporate a {\it truthfulness refocused module} that leverages instruction guidance to further improve truthfulness in contrastive decoding. Extensive experiments on four publicly available datasets demonstrate that the LOL framework significantly  mitigates hallucination while outperforming existing baselines in most cases. For reproducibility, we will release our code and data upon acceptance.
\end{abstract}

\section{Introduction}
By virtue of their extensive pre-training corpus and precise fine-tuning alignment, Large Language Models (LLMs) have demonstrated remarkable capabilities in context-awareness and language-based question-answering \cite{brown2020language, ouyang2022training, schulman2017proximal, radford2019language}. Despite these advancements, LLMs often suffer from a critical issue where their generated content occasionally diverges from user expectations or contradicts real-world knowledge, a phenomenon known as ``Hallucination'' \cite{chuang2023dola}. This issue raises concerns about the accuracy and robustness of LLMs, making hallucination mitigation a pressing challenge that has garnered increasing attention \cite{huang2023survey, zhao2023survey}.

\begin{figure}[t]
	\centering
\includegraphics[width=1\linewidth]{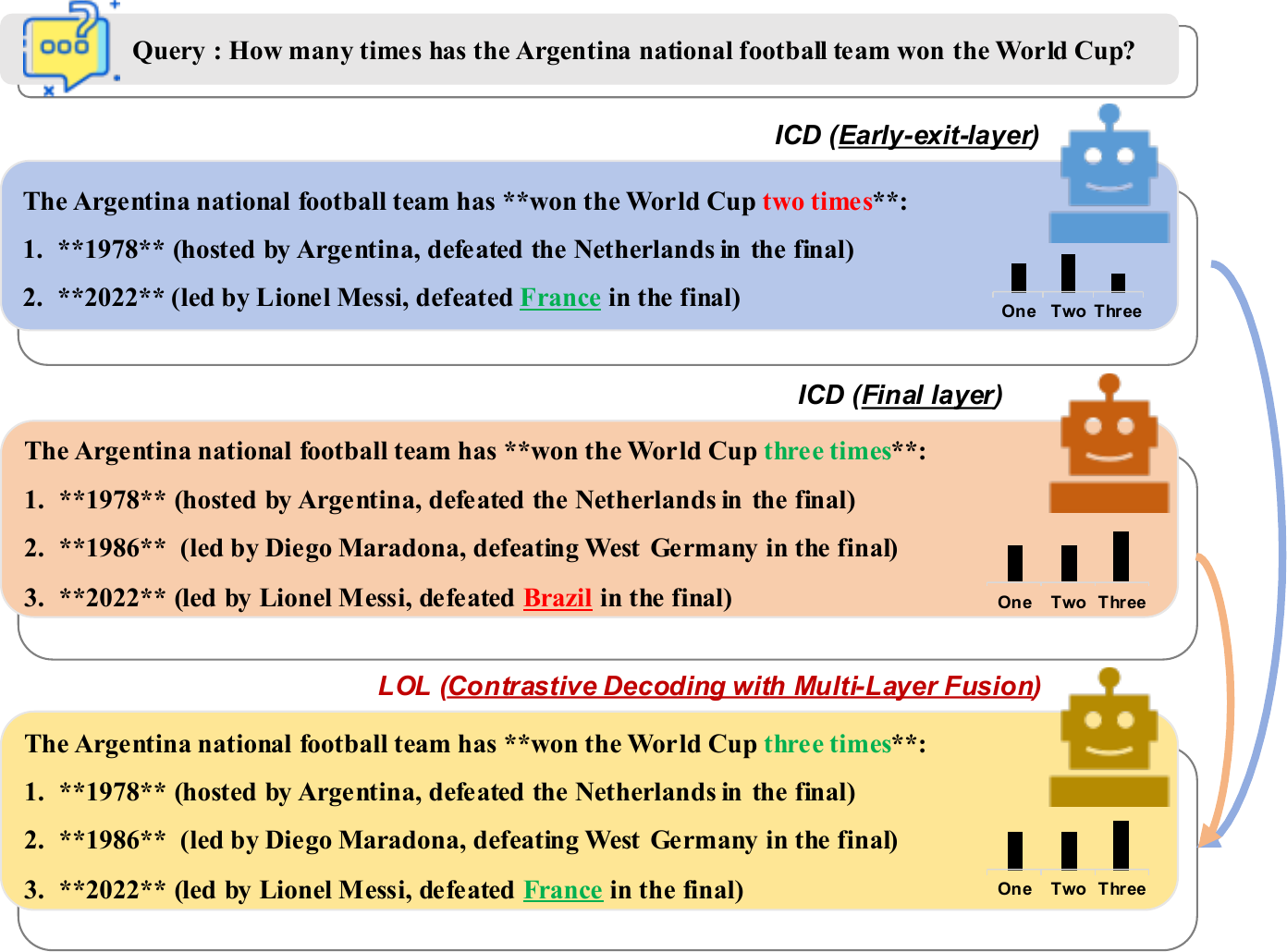}
	\caption{Illustration of mitigating hallucinations with correct knowledge in lower layer.}
	\label{f1}
\end{figure}

Recently,numerous studies have explored hallucination in LLMs. Chuang et al. \cite{chuang2023dola} argue that hallucination stems from imperfections in training and decoding, causing deviations from both training data and factual information in generated content. Similarly, Ye et al. \cite{ye2023cognitive} and Zhang et al. \cite{zhang2023siren} identify pre-training data, model training and optimization, and inference, as the primary sources of hallucination. Specifically, the vast and complex nature of pre-training data inevitably includes erroneous or outdated information, increasing the risk of hallucination. While supervised fine-tuning aligns LLMs and enhances their interactive capabilities, it may also lead to overfitting and cause models to confidently generate answers  beyond their knowledge scope \cite{zhang2023alleviating}. Additionally, the inherent randomness and limitations of decoding strategies can result in inaccurate representations, further contributing to hallucination \cite{huang2023survey, zhang2023siren}.

Existing methods for alleviating hallucination in LLMs vary across different approaches, including reinforcement learning with human feedback \cite{ouyang2022training, stiennon2020learning}, model editing to modify  hallucination knowledge \cite{ilharco2023editing, meng2023mass, ni2023forgetting}, retrieval-augmented generation with leveraging external knowledge bases \cite{peng2023check, wang2023knowledgpt, jiang2023active}, corpus filtering for high-quality data \cite{zhou2024lima}, and representation editing for activation or knowledge refinement \cite{li2023inference, zhang2024truthx}. In this work, we focus on mitigating hallucination during the inference process. Prior studies have explored contrastive decoding, which enhances truthfulness by contrasting the original output distribution with an auxiliary output distribution from another  model or internal layers of the same LLM. This approach effectively refines the original logits by subtracting the logits of a contrasting distribution. For example, 
Zhang et al. \cite{zhang2023alleviating} take a novel approach by inducing hallucination in an amateur model and using it as a contrasting reference for the original model.  This framework, known as ICD, achieves strong results. However, contrastive decoding between the original and amateur model, may distort the original model’s output distribution due to coarse contrast and simplistic subtraction operations. A key challenge is the inherent uncertainty in the amateur model’s training and output, which can undermine the robustness and effectiveness of contrastive decoding.

Chuang et al. \cite{chuang2023dola} propose DoLa and suggest in their experiments that different layers of LLMs may exhibit varying tendencies in knowledge representation. 
They leverage the specific low layer with early exiting found through information theory for contrastive decoding, assuming their outputs lack factual knowledge. 
However, as shown in Figure \ref{f1}, empirically the final layer of LLM sometimes fails to obtain the correct answer or contains incorrect facts, while the middle layer may include needed knowledge, which is lost during transformer processing. In this case, specific early-exit layer is suitable as supplementary knowledge rather than the contrasting distribution to avoid  severe interference.

In this paper, we introduce a novel contrastive decoding framework, \textbf{LOL} (\textbf{\underline{L}}\textbf{\underline{O}}wer \textbf{\underline{L}}ayer Matters). Unlike previous assumptions that lower-layer representations merely lack factual knowledge, We posit that knowledge distributed more intricately across layers 
Instead of using lower-layer outputs purely for contrast, we propose that they can complement and refine the final-layer outputs. 
Specifically, our approach  integrates contrastive decoding at both the final and early-exiting layers between the original and amateur models, enhancing hallucination alleviation. Furthermore, we introduce a plug-and-play truthfulness refocused module that incorporates instruction guidance into contrastive decoding, encouraging LLMs to prioritize factual accuracy during contrastive decoding. Our main contributions are as follows:

\begin{itemize}
\item We propose, \textbf{LOL}, a novel framework for alleviating hallucination, which performs multi-layer fusion contrastive decoding and leverages early-exiting lower layers to refine the final layer outputs.
\item We design a plug-and-play \textbf{truthfulness refocused} module that  enhances factuality accuracy during contrastive decoding.
\item We conduct extensive experiments on four widely used benchmarks, indicating that LOL outperforms existing baselines in most cases.
\end{itemize}

\section{Related Work}

Methods for alleviating hallucination in LLMs are classified into various categories \cite{huang2023survey, zhang2023siren}, including the following:

\textbf{Alignment with assessment feedback} plays a significant role in training LLMs by aligning models with specific feedback while reducing hallucination \cite{stiennon2020learning, liu2023chain, shinn2023reflexion, menick2022teaching}. 
Ouyang et al. \cite{ouyang2022training} propose the RLHF framework, 
which aligns model performance with user intent based on interaction data between humans and LLMs.
Similarly, Shinn et al. \cite{shinn2023reflexion} introduce a method that generates reflective feedback based on previously failed cases to enhance model capabilities.

\textbf{Retrieval-augmented generation} aims to refine LLM outputs by incorporating factual knowledge from external resources, thereby mitigating hallucination \cite{peng2023check, wang2023knowledgpt, li2023large, trivedi2023interleaving, liu2023reta}. Peng et al. \cite{peng2023check} propose an augmentation system 
to enable LLMs to integrate external knowledge for more truthful responses. Trived et al. \cite{trivedi2023interleaving} leverage an improved retrieval method based on the CoT paradigm
to mitigate hallucination in multi-hop question answering. 

\begin{figure*}[t]
	\centering
\includegraphics[width=1\linewidth]{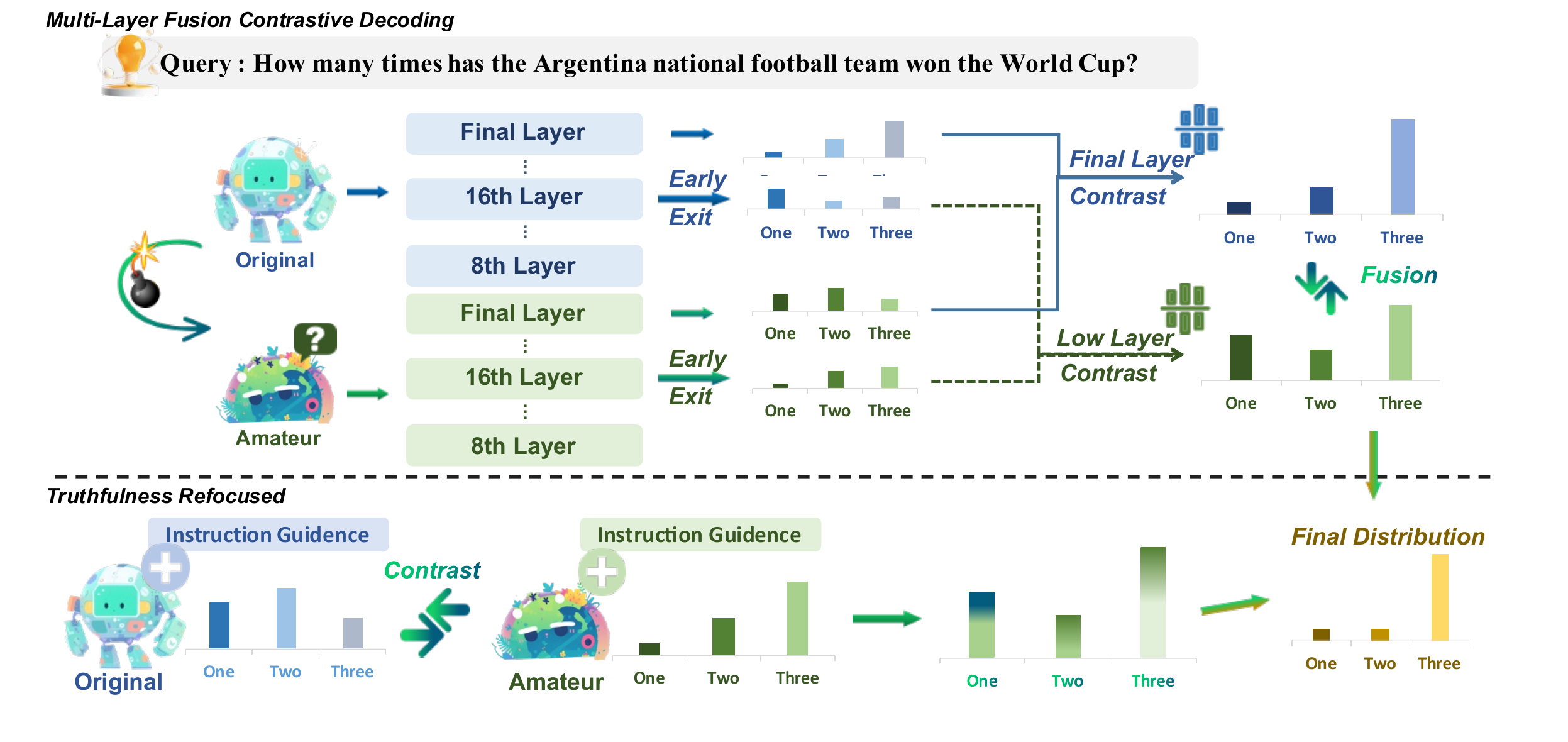}
	\caption{The overview of our proposed \textbf{LOL} framework, incorporating multi-layer fusion and a truthfulness refocused module, built on contrastive decoding.
	}
	\label{f2}
\end{figure*}

\textbf{Model editing} refers to techniques for modifying the knowledge stored in LLMs to correct hallucinated or outdated information 
\cite{zheng2023can, mitchell2022memory, huang2023transformer, meng2022locating, meng2023mass, gupta2023editing}. 
Ni et al. \cite{ni2023forgetting} employ parametric arithmetic to facilitate the forgetting of outdated or hallucinated knowledge 
Mitchell et al. \cite{mitchell2021fast} propose an auxiliary network to help LLMs update their knowledge based on new input-output pairs. For locate-based methods, Meng et al. 
develop MEMIT \cite{meng2023mass}, a novel approach for large scale model editing.

\textbf{Representaion editing} is an inference-time method that modifies and adjusts model output during  inference 
\cite{li2023inference, zhang2024truthx, chen2024truth, burns2022discovering}. Li et al. \cite{li2023inference} propose nference-Time Intervention (ITI), which targets attention heads in LLMs to enhance factuality and alleviate hallucination. Chen et al. \cite{chen2024truth} introduce an orthogonal constraint to improve the identification  of truthful representations. 
Zhang et al. \cite{zhang2024truthx} employ two distinct encoders to separately encode semantic space and truthful spaces, refining factual representations using contrastive learning. 

\textbf{Contrastive decoding} 
operates by contrasting the logits of output distributions between two models---typically one stronger and one weaker in truthfulness---to mitigate hallucination. Li et al. \cite{li2023contrastive} apply contrastive decoding to encourage LLMs to generate both fluent and truthful content. 
Chuang et al. \cite{chuang2023dola} explore the use of specific early layers in an LLM for contrast against the final layer. 
Zhang et al. \cite{zhang2023alleviating} contrast the output of an original model with an amateur model designed to exhibit hallucination, achieving  promising results.

Building on the strengths of contrastive decoding while addressing its limitation, 
we propose a novel framework that extends  contrastive decoding to multiple layers. Our approach first applies contrastive decoding between the final layer and an early-exiting layer of both the original and amateur models. We then merge these outputs to enable multi-layer fusion and introduce a truthfulness refocused module to further improve the factual accuracy.

\section{Method} \label{Method}
In this section, we present our approach to alleviating hallucination in LLMs. Our method consists of two key components: multi-layer fusion contrastive decoding and truthfulness refocusing with instruction guidance, as illustrated in Figure~\ref{f2}. The full implementation is described in Algorithm~1.

\subsection{Multi-Layer Fusion Contrastive Decoding}
To perform contrastive decoding, we first construct an amateur model that exhibits hallucination. Following the setup of \cite{zhang2023alleviating}, we fine-tune the original LLM using a dataset containing non-factual data (e.g., negative samples from fine-tuning datasets) to induce hallucination in the model.

Formally, given an original model $\theta$, we fine-tune it using a hallucination dataset to obtain the amateur model $\theta^{*}$. This procedure is defined as follows:
\begin{equation}\label{key}
    \theta^{*}={\rm FT}\{\theta,D\},
\end{equation}
where FT represents the supervised fine-tuning operation, which can be either full fine-tuning or LoRA fine-tuning. The parameters $\theta$, $\theta^{*}$ denote the original model and the amateur model, respectively, while $D$ is the dataset containing hallucinated or negative samples.

Following the definition of contrastive decoding in \cite{zhang2023alleviating}, given the original model $\theta$ and the amateur model $\theta^{*}$, we apply contrastive decoding by subtracting the output logits of the amateur model $\theta^{*}$ from those of the original model $\theta$ at the \textbf{final layer}. This operation is formally defined as follows:
\begin{equation}
\mathcal{F}_t=\mathrm{log}p(x_t|x_{<t};\theta)-\mathrm{log}p(x_t|x_{<t};\theta^{*}),
\end{equation}
\begin{equation}
p(x_t|x_{<t};\theta)=\mathrm{softmax}(\mathrm{logit}_\theta(x_t|x_{<t})),
\end{equation}
\begin{equation}
p(x_t|x_{<t};\theta^{*})=\mathrm{softmax}(\mathrm{logit}_{\theta^{*}}(x_t|x_{<t})).
\end{equation}

Noting that 
$p(x_t|x_{<t};\theta)$ represents the decoding process of auto-regressive LLMs \cite{zhang2023alleviating, chuang2023dola, li2023contrastive}, where the model predicts the $t-th$ token based on the previous $t-1$ tokens. The term $logit_{\theta}$ denotes the predicted probability distribution.

Similarly, we perform a parallel operation to obtain the contrastive decoding result from a \textbf{lower layer with early exiting}, defined as follows:
\begin{equation}
\mathcal{F}_t^{'}=\mathrm{log} p(x_t|x_{<t};\theta;L)-\mathrm{log}p(x_t|x_{<t};\theta^{*};L),
\end{equation}

where 
$L$ represents the specific layer chosen for early exiting in contrastive decoding. We control the hyperparameter $L$ to select an appropriate layer, ensuring a balance between early-exiting contrastive decoding and final-layer contrastive decoding. The purpose of this module is to leverage the lower-layer contrastive decoding results to refine the final-layer contrastive decoding. This helps mitigate potential issues arising from naïve contrast subtraction, which may introduce excessive perturbation in the probability distribution. Unlike  \cite{chuang2023dola}, where lower layers are considered weaker counterparts for contrastive decoding, we 
empirically consider them as complementary to the final layer, providing a balance rather than merely serving as a contrasting reference.
To achieve multi-layer fusion, we merge the results from both contrastive decoding operations as follows:
\begin{equation}
\mathcal{F}_{ML}=\mathcal{F}_t + \omega\mathcal{F}^{'}_t,
\end{equation}

where $\mathcal{F}_{ML}$ denotes the result of our multi-layer fusion contrastive decoding, and $\omega \in (0,1]$ is a hyperparameter that controls the contribution of the lower-layer contrastive decoding result.
\subsection{Truthfulness Refocused}
To further enhance truthfulness in contrastive decoding, we introduce a plug-and-play module called truthfulness refocused. This module encourages LLMs to focus on the the factuality of generated content, using instruction guidance. The objective is to highlight  factual words and enhance more the granularity of contrastive decoding.

Specifically, we introduce an instruction guidance $x_{instruction}$ into the prefix instruction, guiding the model to focus on key words in the generated content that contribute to the overall factuality of the sentence. This process is defined as follows:
\begin{equation}
\begin{aligned}
\mathcal{F}_{TR}&=\mathrm{log}p(x_t|(x_{<t}||x_{instruction});\theta)  \\
&-\mathrm{log}p(x_t|(x_{<t}||x_{instruction});\theta^{*}),
\end{aligned}
\end{equation}

where $\mathcal{F}_{TR}$ refers to the result of contrastive decoding from the truthfulness refocused module.

Finally, we combine the two module of our framework to obtain the final logits of the output distribution, which are then used for prediction and generation, as follows:
\begin{equation}
\mathcal{F}_{Final}=\mathcal{F}_{ML} + \omega^{'}\mathcal{F}_{TR}
\end{equation}
\begin{equation}
p(x_t|x_{<t})=\mathrm{softmax}(\mathcal{F}_{Final}),
\end{equation}
where $\mathcal{F}_{Final}$ refers to the result of our LOL framework based on contrastive decoding. $\omega^{'} \in (0,1]$ is a hyperparameter controlling the fusion  ratio. We apply the $softmax$ function to regularize the logits output and derive the final result.

\section{Experiment}

\subsection{Datasets}
We use  four widely recognized benchmarks for evaluating hallucination alleviation and logical reasoning: \textbf{TruthfulQA} \cite{lin2022truthfulqa}, \textbf{FACTOR} \cite{muhlgay2024generating}, \textbf{StrategyQA} \cite{geva2021strategyqa}, and \textbf{GSM8K} \cite{cobbe2021gsm8k}. 

\textbf{TruthfulQA}, a well-known dataset for assessing the truthfulness of LLMs,  consists of 817 questions across 38 categories. We  conduct experiments using the multiple-choice format, where the task involves identifying and scoring  different answers from a set of best answers, correct answers, and incorrect answers. We evaluate the results using three metrics: \textbf{MC1}, \textbf{MC2}, and \textbf{MC3}. Specifically, \textbf{MC1} assesses whether the model assigns the highest probability to the best answer among all choices; \textbf{MC2} checks if the normalized probability of all correct answers (including the best answer) is higher than that of all incorrect answers; and \textbf{MC3} evaluates whether the probability  of each correct answer is greater than that of all incorrect answers. For the \textbf{FACTOR} dataset, which focuses on content completion. We conduct experiments on its three sub-datasets: (\textbf{News}, \textbf{Wiki}, and \textbf{Expert}). We use accuracy as the evaluation metric for text completion accuracy. Furthermore, we use \textbf{StrategyQA} and \textbf{GSM8K} to test the effectiveness of our approach on generation tasks and reasoning. StrategyQA consists of 2,288 logic-based question-answer pairs, while GSM8K contains over 8,000 high-quality math problems for evaluating mathematical reasoning. Similarly, we use accuracy as the primary evaluation metric for both.

What requires special attention is that, following the approach in \cite{zhang2023alleviating}, we use \textbf{HaluEval} \cite{li2023halueval} to construct our amateur model with induced hallucination for contrastive decoding. HaluEval is a well-known dataset that contains 35,000 negative samples with hallucination. It consists of question-answer pairs generated by ChatGPT, along with instructions for a variety of downstream tasks. 

\begin{table*}[t] \small
\setlength\tabcolsep{11pt}
\renewcommand\arraystretch{1.2}
\begin{center}
\begin{tabular}{lcccccccc}

\toprule
\multirow{2}{*}{\textbf{Method}}&\multicolumn{3}{c}{\textbf{TruthfulQA}}&\multicolumn{3}{c}{\textbf{FACTOR}} &\multicolumn{2}{c}{\textbf{CoT}} \\
\cmidrule(r){2-4}  \cmidrule(r){5-7} \cmidrule(r){8-9}
& \textbf{MC1} & \textbf{MC2} & \textbf{MC3} & \textbf{News} & \textbf{Wiki} & \textbf{Expert} & \textbf{StrQA} & \textbf{GSM8K}\\
\hline 
\rowcolor{gray!20}\multicolumn{9}{c}{\textbf{LLAMA3-8B-Instruct}} \\
Greedy  & 43.13	&61.26	&33.89	&70.44	&59.15	&63.78&	\underline{72.67}&	75.78\\

ITI & 43.39	&61.53	&33.94	&60.19	&47.22	&52.76&	68.17&69.20\\

SH2 & 43.30&	64.47&	36.23&	70.80&	59.69&	64.22& 72.38	&76.92\\

DoLa & 42.96&65.76&35.71&70.10&59.37&64.06&72.22&76.30\\

ICD & \underline{61.76}	&\underline{79.63}	&\underline{58.90}	&\underline{71.99}	&\underline{60.55}	&\textbf{65.51}&	72.45	&\underline{77.35}\\

\textbf{LOL (Ours)} & \textbf{64.50}	&\textbf{81.12}	&\textbf{62.25}	&\textbf{72.60}&	\textbf{62.04}	&\underline{65.44}	&\textbf{73.74}	&\textbf{78.30}\\

\rowcolor{gray!20}\multicolumn{9}{c}{\textbf{Mistral-7B-Instruct-v0.2}} \\

Greedy  & 55.26&72.08&44.33&74.50&60.38&65.51&67.87&43.09\\

ITI & 55.12	&72.30	&44.87	&68.76	&56.74	&59.82&	62.33	&39.70\\

SH2 & 54.29	&75.32	&47.80	&74.93	&\underline{61.02}	&67.40&	67.54	&\underline{43.37}\\

DoLa & 52.19&77.85&48.21&73.82&60.62&66.70&67.74&42.33\\

ICD & \underline{62.62}	&\underline{79.83}	&\underline{56.37}	&\underline{75.76}	&60.92	&\underline{68.95}&	\underline{68.17}	&42.07\\

\textbf{LOL (Ours)} &\textbf{63.50}	&\textbf{80.40}	&\textbf{57.94}	&\textbf{76.44}&	\textbf{61.67}	&\textbf{69.12}	&\textbf{68.94}	&\textbf{43.65}\\

\rowcolor{gray!20}\multicolumn{9}{c}{\textbf{LLAMA2-7B-Chat}} \\
Greedy  & 37.00&54.65	&27.82	&64.71	&56.61	&64.85&	63.67&	21.64\\

ITI & 37.01	&54.66	&27.82	&53.28	&43.82	&51.69&	58.74	&17.86\\

SH2 & 33.90	&57.07	&29.79	&\underline{65.31}	&57.37	&67.22&	64.40 &\underline{22.17}\\

CD & 28.15	&54.87	&29.75	&64.57	&\textbf{58.47}	&67.12	&58.42	&15.04\\

DoLa & 32.97&	60.84&	29.50&	64.32&	\underline{57.63}&	67.30&	64.16&	22.07\\

ICD & \underline{45.09}	&\underline{69.10}	&\underline{41.59}	&65.20	&56.57	&\underline{67.66}&	\underline{64.47}	&21.72\\

\textbf{LOL (Ours)} & \textbf{49.87}	&\textbf{73.62}	&\textbf{46.53}	&\textbf{65.96}&	57.14	&\textbf{71.11}	&\textbf{65.81}	&\textbf{22.56}\\

\bottomrule

\end{tabular}
\end{center}
\caption{Experimental results on LLAMA3-8B-Instruct, Mistral-7B-Instruct-v0.2, and LLAMA2-7B-Chat across the TruthfulQA, FACTOR, StrategyQA (StrQA), and GSM8K datasets. The best and second-best results are \textbf{bolded} and \underline{underlined}, respectively. Our method outperforms consistently other baselines across all four benchmarks.}
\label{tab1}
\end{table*}

\subsection{Baselines}

To evaluate the effectiveness of our proposed LOL method for mitigating hallucination, we conduct experiments using the following inference-time baselines: \\
\textbf{Greedy decoding}: A common decoding strategy for auto-regressive LLMs, which selects the token with the highest probability for prediction and generation. \\
\textbf{CD} \cite{li2023contrastive}: An initial and straightforward method of contrastive decoding, which performs contrastive decoding between two different scales of LLMs with different parameters. Due to model scale limitations, we follow the settings of \cite{zhang2023alleviating, zhang2024truthx} and perform contrastive decoding between the 13B-Chat and 7B-Chat models for the contrastive decoding experiments on LLAMA2-7B-Chat. \\
\textbf{ITI} \cite{li2023inference}: \underline{I}nference-\underline{T}ime-\underline{I}ntervention is a representation editing method that alleviates hallucination by editing the model's activation for attention heads during inference stage.\\
\textbf{DoLa} \cite{chuang2023dola}: An improved method based on contrastive decoding, which determines the low layers with early exiting as the target for contrast with the final layer by maximizing the Jensen-Shannon divergence.\\ \textbf{SH2} \cite{kai2024sh2}: This method focuses on low-confidence tokens during inference, using a re-think mechanism to enhance truthfulness.\\
\textbf{ICD} \cite{zhang2023alleviating}: An effective method that performs contrastive decoding between the original model and an amateur model induced with hallucination, yielding promising results.

\subsection{Experiments Results}

The experimental results, as shown in Table \ref{tab1}, demonstrate that our proposed LOL framework consistently outperforms other baselines across multiple models and benchmarks. The results highlight the effectiveness of our method in alleviating hallucination,  achieved through the combination of multi-layer fusion contrastive decoding and the truthfulness refocused mechanism. More details regarding the completion details and the selection of hyperparameters are provided in Section §\ref{hyperparameters}. 

Specifically, on TruthfulQA, LOL shows significant improvements, surpassing the best baseline (ICD) by $+2.74$, $+1.49$, and $+3.35$ points on LLAMA3-8B-Instruct. Additionally, on LLAMA2-7B-Chat, LOL achieves an average improvement of $4.75$ points across three indicators, demonstrating the robustness of our approach even with smaller LLMs. On the FACTOR dataset, LOL delivers small but consistent  improvements across all three subsets, proving its effectiveness across various content domains. However, task-specific nuances, such as external knowledge dominance, may affect its performance in some cases. For generation tasks, LOL demonstrates a clear advantage, improving by 1.13 points on StrategyQA and 1.12 points on GSM8K on average across the three base models. This highlights the adaptability of LOL, not only in enhancing factuality but also in improving logical and mathematical reasoning capabilities.

Moreover, the ITI baseline shows a considerable decline in several metrics, likely due to its representation editing of model activations, which may disrupt the model's output distribution. The CD baseline, although effective in some areas, experiences a notable drop in the MC1 metric on TruthfulQA, potentially due to the contrastive decoding between two models of similar types but only slight parameter scale differences. While DoLa, which contrasts the lower and final layers, sees significant improvements due to the divergence between layers, ICD performs contrastive decoding between the original and amateur models. Our LOL framework outperforms ICD across most metrics, emphasizing the efficacy of the multi-layer fusion and truthfulness refocused modules.

\section{Analysis}
\subsection{Ablation Study}
To assess the individual contributions of the multi-layer fusion and truthfulness refocused modules in the LOL framework, we conduct an ablation study on the aforementioned benchmarks. In this study, we remove each module separately: the multi-layer fusion module (denoted as \textbf{``w/o Multi-Layer Fusion''}) and the truthfulness refocused module (denoted as \textbf{``w/o Truthfulness Refocused''}), respectively. The ablation results are summarized in Table \ref{tab2}, with the best and second-best performances \textbf{bolded} and \underline{underlined} for emphasis. The results indicate that both modules contribute to enhancing LLM performance in most cases, outperforming the ICD baseline. Notably, the multi-layer fusion module plays a more significant role in improving performance compared to the truthfulness refocused module. This highlights the importance of multi-layer fusion in our framework for mitigating hallucinations effectively.

\begin{table*}[t] \footnotesize
\setlength\tabcolsep{8pt}
\renewcommand\arraystretch{1.1}
\begin{center}
\begin{tabular}{lcccccccc}

\toprule
\multirow{2}{*}{\textbf{Method}}&\multicolumn{3}{c}{\textbf{TruthfulQA}}&\multicolumn{3}{c}{\textbf{FACTOR}} &\multicolumn{2}{c}{\textbf{CoT}} \\
\cmidrule(r){2-4}  \cmidrule(r){5-7} \cmidrule(r){8-9}
& \textbf{MC1} & \textbf{MC2} & \textbf{MC3} & \textbf{News} & \textbf{Wiki} & \textbf{Expert} & \textbf{StrQA} & \textbf{GSM8K}\\
\hline 
\rowcolor{gray!20}\multicolumn{9}{c}{\textbf{LLAMA3-8B-Instruct}} \\

ICD & 61.76	&79.63	&58.90&71.99&60.55&\underline{65.51}&	72.45	&77.35\\

\textbf{LOL (Ours)} & \textbf{64.50}	&\textbf{81.12}	&\textbf{62.25}	&\textbf{72.60}&	\textbf{62.04}	&65.44	&\textbf{73.74}	&\textbf{78.30}\\

\textit{w/o} Multi-Layer Fusion &62.12 & 80.03 & 59.64 & \underline{72.19} & 60.88 & 65.23  & 72.95 & 77.52 \\

\textit{w/o} Truthfulness Refocused &\underline{63.84}	&\underline{80.60}&	\underline{60.74}&71.89&\underline{61.25}&\textbf{65.94}&\underline{73.39}&\underline{77.88}\\

\rowcolor{gray!20}\multicolumn{9}{c}{\textbf{Mistral-7B-Instruct-v0.2}} \\

ICD & 62.62	&79.83	&56.37	&75.76	&\underline{60.92}	&68.95&68.17	&42.07\\

\textbf{LOL (Ours)} &\textbf{63.50}	&\textbf{80.40}	&\textbf{57.94}	&\textbf{76.44}&	\textbf{61.67}	&\underline{69.12}	&\textbf{68.94}	&\textbf{43.65}\\

\textit{w/o} Multi-Layer Fusion & 62.27 & 79.61 & 56.69 & 75.32 & 60.59 & 68.43 & \underline{68.59} & 42.77\\

\textit{w/o} Truthfulness Refocused &\underline{62.99}&\underline{79.96}&\underline{56.81}&\underline{75.96}&60.85&\textbf{69.39}&68.30&\underline{43.21}\\

\bottomrule

\end{tabular}
\end{center}
\caption{Ablation study on LLAMA3-8B-Instruct and Mistral-7B-Instruct-v0.2 across four benchmarks. The best and second-best results are \textbf{bolded} and \underline{underlined}, respectively.}
\label{tab2}
\end{table*}

\begin{table}[t] \footnotesize
\setlength\tabcolsep{11pt}
\renewcommand\arraystretch{1.1}
\begin{center}
\begin{tabular}{lccc}
\toprule
\multirow{2}{*}{\textbf{Method}}&\multicolumn{3}{c}{\textbf{TruthfulQA}} \\
\cline{2-4}  
& \textbf{MC1} & \textbf{MC2} & \textbf{MC3}  \\
\hline
\rowcolor{gray!20}\multicolumn{4}{c}{\textbf{LLAMA2-13B-Chat}} \\
Greedy & 37.75 &55.67 &28.16\\

ICD & 48.47 &73.47 &46.04\\

\textbf{LOL} (7B-Chat) & \textbf{49.87} &\textbf{73.62} &\textbf{46.53} \\

\rowcolor{gray!20}\multicolumn{4}{c}{\textbf{Mistral-7B-Instruct-v0.1}} \\
Greedy & 39.09 &55.80 &28.25\\

ICD & 58.53 &74.73 &50.38 \\

\textbf{LOL} & \textbf{59.83}	&\textbf{75.20}	&\textbf{51.62} \\

\rowcolor{gray!20}\multicolumn{4}{c}{\textbf{Baichuan2-7B-Chat}} \\
Greedy & 34.93 &52.14 &27.19\\

+ICD & 45.75 &65.51 &39.67 \\

\textbf{LOL} & \textbf{48.11}	&\textbf{68.30}	&\textbf{42.31} \\

\bottomrule

\end{tabular}
\caption{Experimental results of adaptability testing with three model settings on TruthfulQA.}
\label{tab3}
\end{center}
\end{table}

\subsection{Adaptability Testing}
\begin{figure*}[t]
	\centering
\includegraphics[width=1\linewidth]{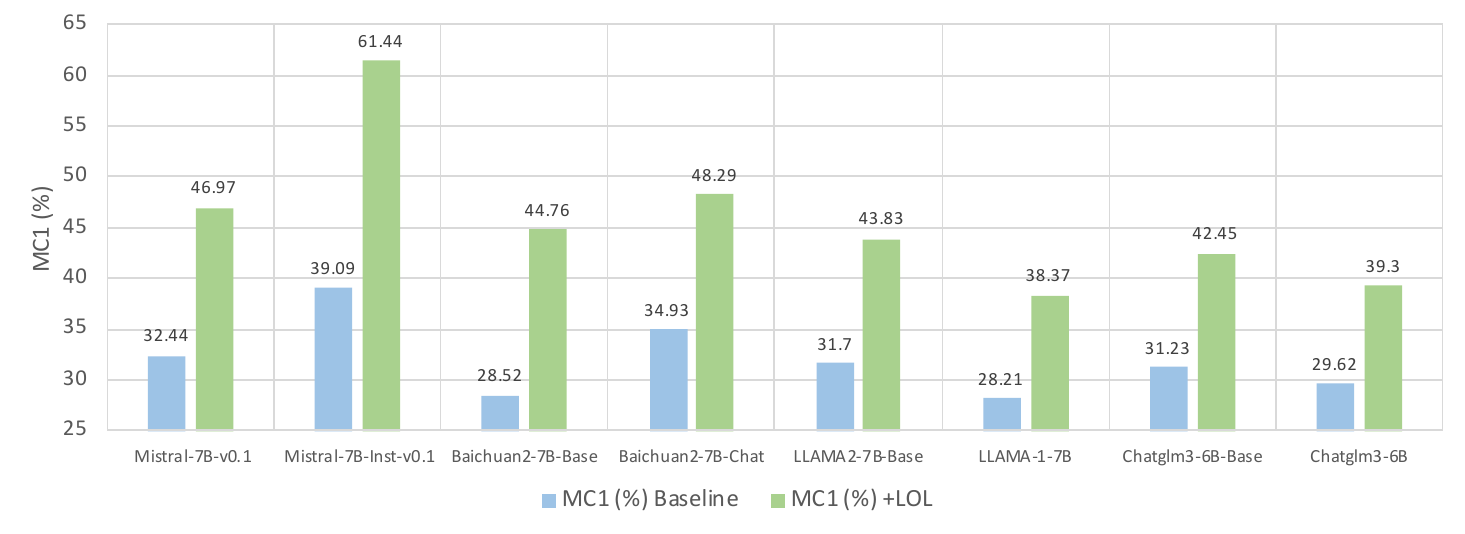}
	\caption{Experimental results on TruthfulQA (MC1 \%) across eight different base models.}
	\label{f3}
\end{figure*}

We conduct experiments on different model scales and architectures to evaluate the adaptability of our method. Particularly, we apply LOL to eight base LLMs on the TruthfulQA benchmark, as shown in Figure~\ref{f3}. Experimental results show that LOL effectively alleviates hallucination across diverse model architectures and scales. Additionally, the results highlight that the HaluEval dataset provides universal hallucination patterns consistent with real world scenarios, making it valuable resource for building a generalizable amateur model.

We conduct an in-depth comparison between our LOL method and the ICD method \cite{zhang2023alleviating} on TruthfulQA, using LLAMA2-Chat, Baichuan2-7B-Chat, and Mistral-7B-Instruct-v0.1 as base models. The experimental results are presented in Table~\ref{tab3}. Notably, despite using LLAMA2-7B as the base model, our method outperforms the ICD method, which is based on the larger LLAMA2-13B model. Furthermore, results from different model families demonstrate that our method generalizes well across various LLM architectures, highlighting its strong adaptability.

\subsection{Analysis of Fusion with Different Layers}
\begin{figure}[t]
	\centering
\includegraphics[width=0.9\linewidth]{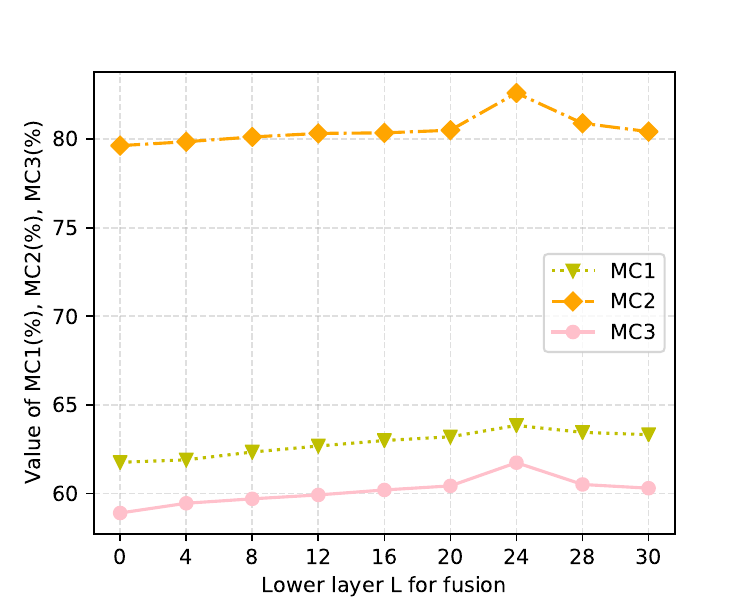}
	\caption{Experimental results of multi-layer fusion contrastive decoding with different lower layers on TruthfulQA using LLAMA3-8B-Instruct.}
	\label{f4}
\end{figure}

\begin{figure}[t]
	\centering
\includegraphics[width=0.9\linewidth]{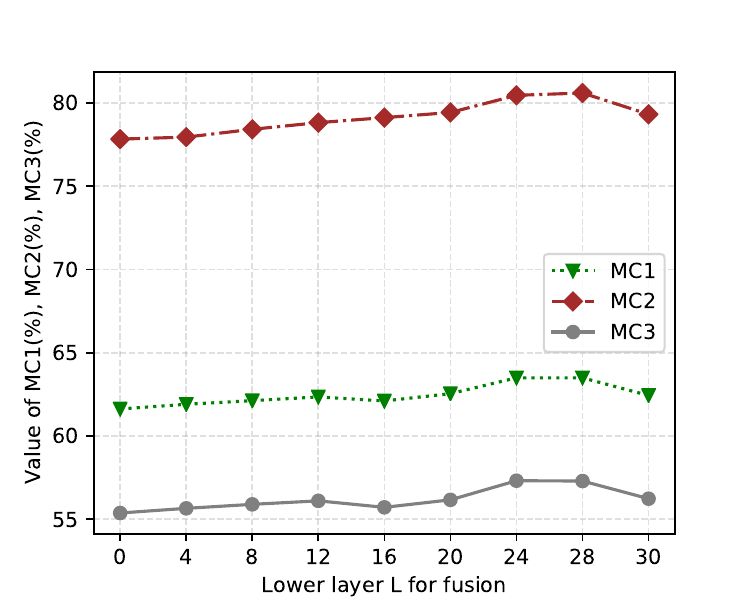}
	\caption{Experimental results of multi-layer fusion contrastive decoding with different low layers on TruthfulQA using Mistral-7B-Instruct-v0.2.}
	\label{f5}
\end{figure}

Our proposed method for alleviating hallucination introduces contrastive decoding between lower layers with early exists within the original and amateur models, enabling multi-layer fusion contrastive decoding. To assess its robustness and interpretability, we analyze the impact of different low-layer selections. The results are presented in Figures~\ref{f4}~and~\ref{f5}. Specifically, we conduct experiments on TruthfulQA using LLAMA3-8B-Instruct and Mistral-7B-Instruct-v0.2 alongside their corresponding amateur models. For clarity, We exclude the truthfulness refocused module and evaluate performance across selected layers $L=[4,8,12,16,20,24,28,30]$. 

We take $L=0$ as the baseline, representing contrastive decoding without multi-layer fusion (i.e., ICD \cite{zhang2023alleviating}). Overall, our method benefits from the multi-layer fusion module in most cases. Notably, contrastive decoding with certain higher layers (such as the $24th$ and $28th$ layers) proves to be more effective in mitigating hallucination, whereas the effect diminishes when using significantly higher or lower layers. This aligns with the role of different layers within the model. For instance, experimental results on both models indicate that the $24th$ layer may contain supplementary knowledge that enhances  the final layer's performance. Understanding the distribution of knowledge across layers is  crucial for further exploring LLM interpretability. In §\ref{case study}, we will examine the impact of specific lower layers on the generation task through a detailed case study.

\subsection{Analysis of Strength for Truthfulness
Refocused}
To evaluate the impact of the truthfulness refocused module on our proposed approach, we conduct a parameter sensitivity analysis. Specifically, we vary the strength of truthfulness refocusing by adjusting the hyperparameter $\omega^{'}$. The results, presented in Figure~\ref{f11} (in Appendix), are based on experiment conducted on the TruthfulQA benchmark using LLAMA3-8B-Instruct. Notably, we retain the multi-layer fusion module to assess the compatibility of both components. We examine five different values of $\omega^{'}=[0.1,0.3,0.5,0.7,1.0]$. Additionally, we set the lower-layer early exiting parameter to 24, as previous experiments indicate that this configuration yields the best performance.

We take $\omega^{'}=0$ as the results of contrastive decoding without the truthfulness refocused module. The experimental results reveal that the truthfulness refocused module we designed leads to overall improvement. The best results are obtained when $\omega^{'}$ is set to $0.5$ or $1.0$.Overall, the effect of our method remains relatively stable as $\omega^{'}$ changes, demonstrating the robustness and stability of this module.

\section{Conclusion}

In this paper, we introduce a novel framework for alleviating hallucination in LLMs, called \textbf{LOL} (\textbf{\underline{L}}\textbf{\underline{O}}wer \textbf{\underline{L}}ayer Matters). LOL utilizes contrastive decoding between lower layers and early exits, facilitating multi-layer fusion contrastive decoding with the final layer. In addition, we propose a truthfulness refocused module that further enhances content accuracy during the decoding process. Our extensive  experiments across various benchmarks show that LOL consistently outperforms  other baselines, highlighting its effectiveness. Further analysis supports the robustness and adaptability of LOL, affirming its potential across different model architectures. Future work will focus on refining the efficiency and effectiveness of contrastive decoding to further mitigate hallucination.

\section{Limitations}

While the \textbf{LOL} framework significantly alleviates hallucination by utilizing multi-layer fusion contrastive decoding and refining outputs through low-layer early-exits, it does introduce additional computational overhead. This arises from the simultaneous use of multiple logits distributions during decoding, which may increase resource requirements and processing time.

\bibliography{anthology,custom}

\begin{thebibliography}{45}
\expandafter\ifx\csname natexlab\endcsname\relax\def\natexlab#1{#1}\fi

\bibitem[{AI@Meta(2024)}]{llama3modelcard}
AI@Meta. 2024.
\newblock \href {https://github.com/meta-llama/llama3/blob/main/MODEL_CARD.md} {Llama 3 model card}.

\bibitem[{Brown et~al.(2020)Brown, Mann, Ryder, Subbiah, Kaplan, Dhariwal, Neelakantan, Shyam, Sastry, Askell et~al.}]{brown2020language}
Tom Brown, Benjamin Mann, Nick Ryder, Melanie Subbiah, Jared~D Kaplan, Prafulla Dhariwal, Arvind Neelakantan, Pranav Shyam, Girish Sastry, Amanda Askell, et~al. 2020.
\newblock Language models are few-shot learners.
\newblock \emph{Advances in neural information processing systems}, 33:1877--1901.

\bibitem[{Burns et~al.(2022)Burns, Ye, Klein, and Steinhardt}]{burns2022discovering}
Collin Burns, Haotian Ye, Dan Klein, and Jacob Steinhardt. 2022.
\newblock Discovering latent knowledge in language models without supervision.
\newblock In \emph{The Eleventh International Conference on Learning Representations}.

\bibitem[{Chen et~al.(2024)Chen, Sun, Jiao, Lian, Kang, Wang, and Xu}]{chen2024truth}
Zhongzhi Chen, Xingwu Sun, Xianfeng Jiao, Fengzong Lian, Zhanhui Kang, Di~Wang, and Chengzhong Xu. 2024.
\newblock Truth forest: Toward multi-scale truthfulness in large language models through intervention without tuning.
\newblock In \emph{Proceedings of the AAAI Conference on Artificial Intelligence}, volume~38, pages 20967--20974.

\bibitem[{Chuang et~al.(2023)Chuang, Xie, Luo, Kim, Glass, and He}]{chuang2023dola}
Yung-Sung Chuang, Yujia Xie, Hongyin Luo, Yoon Kim, James~R Glass, and Pengcheng He. 2023.
\newblock Dola: Decoding by contrasting layers improves factuality in large language models.
\newblock In \emph{The Twelfth International Conference on Learning Representations}.

\bibitem[{Cobbe et~al.(2021)Cobbe, Kosaraju, Bavarian, Chen, Jun, Kaiser, Plappert, Tworek, Hilton, Nakano, Hesse, and Schulman}]{cobbe2021gsm8k}
Karl Cobbe, Vineet Kosaraju, Mohammad Bavarian, Mark Chen, Heewoo Jun, Lukasz Kaiser, Matthias Plappert, Jerry Tworek, Jacob Hilton, Reiichiro Nakano, Christopher Hesse, and John Schulman. 2021.
\newblock Training verifiers to solve math word problems.
\newblock \emph{arXiv preprint arXiv:2110.14168}.

\bibitem[{Geva et~al.(2021)Geva, Khashabi, Segal, Khot, Roth, and Berant}]{geva2021strategyqa}
Mor Geva, Daniel Khashabi, Elad Segal, Tushar Khot, Dan Roth, and Jonathan Berant. 2021.
\newblock {Did Aristotle Use a Laptop? A Question Answering Benchmark with Implicit Reasoning Strategies}.
\newblock \emph{Transactions of the Association for Computational Linguistics (TACL)}.

\bibitem[{Gupta et~al.(2023)Gupta, Mondal, Sheshadri, Zhao, Li, Wiegreffe, and Tandon}]{gupta2023editing}
Anshita Gupta, Debanjan Mondal, Akshay Sheshadri, Wenlong Zhao, Xiang Li, Sarah Wiegreffe, and Niket Tandon. 2023.
\newblock Editing common sense in transformers.
\newblock In \emph{Proceedings of the 2023 Conference on Empirical Methods in Natural Language Processing}, pages 8214--8232.

\bibitem[{Hu et~al.(2021)Hu, Wallis, Allen-Zhu, Li, Wang, Wang, Chen et~al.}]{hu2021lora}
Edward~J Hu, Phillip Wallis, Zeyuan Allen-Zhu, Yuanzhi Li, Shean Wang, Lu~Wang, Weizhu Chen, et~al. 2021.
\newblock Lora: Low-rank adaptation of large language models.
\newblock In \emph{International Conference on Learning Representations}.

\bibitem[{Huang et~al.(2023{\natexlab{a}})Huang, Yu, Ma, Zhong, Feng, Wang, Chen, Peng, Feng, Qin et~al.}]{huang2023survey}
Lei Huang, Weijiang Yu, Weitao Ma, Weihong Zhong, Zhangyin Feng, Haotian Wang, Qianglong Chen, Weihua Peng, Xiaocheng Feng, Bing Qin, et~al. 2023{\natexlab{a}}.
\newblock A survey on hallucination in large language models: Principles, taxonomy, challenges, and open questions.
\newblock \emph{arXiv preprint arXiv:2311.05232}.

\bibitem[{Huang et~al.(2023{\natexlab{b}})Huang, Shen, Zhang, Zhou, Rong, and Xiong}]{huang2023transformer}
Zeyu Huang, Yikang Shen, Xiaofeng Zhang, Jie Zhou, Wenge Rong, and Zhang Xiong. 2023{\natexlab{b}}.
\newblock Transformer-patcher: One mistake worth one neuron.
\newblock In \emph{The Eleventh International Conference on Learning Representations}.

\bibitem[{Ilharco et~al.(2023)Ilharco, Ribeiro, Wortsman, Schmidt, Hajishirzi, and Farhadi}]{ilharco2023editing}
Gabriel Ilharco, Marco~Tulio Ribeiro, Mitchell Wortsman, Ludwig Schmidt, Hannaneh Hajishirzi, and Ali Farhadi. 2023.
\newblock Editing models with task arithmetic.
\newblock In \emph{The Eleventh International Conference on Learning Representations}.

\bibitem[{Jiang et~al.(2023)Jiang, Xu, Gao, Sun, Liu, Dwivedi-Yu, Yang, Callan, and Neubig}]{jiang2023active}
Zhengbao Jiang, Frank~F Xu, Luyu Gao, Zhiqing Sun, Qian Liu, Jane Dwivedi-Yu, Yiming Yang, Jamie Callan, and Graham Neubig. 2023.
\newblock Active retrieval augmented generation.
\newblock \emph{arXiv preprint arXiv:2305.06983}.

\bibitem[{Kai et~al.(2024)Kai, Zhang, Hu, and Lin}]{kai2024sh2}
Jushi Kai, Tianhang Zhang, Hai Hu, and Zhouhan Lin. 2024.
\newblock Sh2: Self-highlighted hesitation helps you decode more truthfully.
\newblock \emph{arXiv preprint arXiv:2401.05930}.

\bibitem[{Li et~al.(2023{\natexlab{a}})Li, Rawat, Zaheer, Wang, Lukasik, Veit, Yu, and Kumar}]{li2023large}
Daliang Li, Ankit~Singh Rawat, Manzil Zaheer, Xin Wang, Michal Lukasik, Andreas Veit, Felix Yu, and Sanjiv Kumar. 2023{\natexlab{a}}.
\newblock Large language models with controllable working memory.
\newblock In \emph{Findings of the Association for Computational Linguistics: ACL 2023}, pages 1774--1793.

\bibitem[{Li et~al.(2023{\natexlab{b}})Li, Cheng, Zhao, Nie, and Wen}]{li2023halueval}
Junyi Li, Xiaoxue Cheng, Wayne~Xin Zhao, Jian-Yun Nie, and Ji-Rong Wen. 2023{\natexlab{b}}.
\newblock Halueval: A large-scale hallucination evaluation benchmark for large language models.
\newblock In \emph{Proceedings of the 2023 Conference on Empirical Methods in Natural Language Processing}, pages 6449--6464.

\bibitem[{Li et~al.(2023{\natexlab{c}})Li, Patel, Vi{\'e}gas, Pfister, and Wattenberg}]{li2023inference}
Kenneth Li, Oam Patel, Fernanda Vi{\'e}gas, Hanspeter Pfister, and Martin Wattenberg. 2023{\natexlab{c}}.
\newblock Inference-time intervention: Eliciting truthful answers from a language model.
\newblock \emph{arXiv preprint arXiv:2306.03341}.

\bibitem[{Li et~al.(2023{\natexlab{d}})Li, Holtzman, Fried, Liang, Eisner, Hashimoto, Zettlemoyer, and Lewis}]{li2023contrastive}
Xiang~Lisa Li, Ari Holtzman, Daniel Fried, Percy Liang, Jason Eisner, Tatsunori~B Hashimoto, Luke Zettlemoyer, and Mike Lewis. 2023{\natexlab{d}}.
\newblock Contrastive decoding: Open-ended text generation as optimization.
\newblock In \emph{Proceedings of the 61st Annual Meeting of the Association for Computational Linguistics (Volume 1: Long Papers)}, pages 12286--12312.

\bibitem[{Lin et~al.(2022)Lin, Hilton, and Evans}]{lin2022truthfulqa}
Stephanie Lin, Jacob Hilton, and Owain Evans. 2022.
\newblock Truthfulqa: Measuring how models mimic human falsehoods.
\newblock In \emph{Proceedings of the 60th Annual Meeting of the Association for Computational Linguistics (Volume 1: Long Papers)}, pages 3214--3252.

\bibitem[{Liu et~al.(2023{\natexlab{a}})Liu, Sferrazza, and Abbeel}]{liu2023chain}
Hao Liu, Carmelo Sferrazza, and Pieter Abbeel. 2023{\natexlab{a}}.
\newblock Chain of hindsight aligns language models with feedback.
\newblock In \emph{The Twelfth International Conference on Learning Representations}.

\bibitem[{Liu et~al.(2023{\natexlab{b}})Liu, Jin, Wang, Cheng, Dou, and Wen}]{liu2023reta}
J~Liu, J~Jin, Z~Wang, J~Cheng, Z~Dou, and J~Wen. 2023{\natexlab{b}}.
\newblock Reta-llm: A retrieval-augmented large language model toolkit. arxiv, abs/2306.05212.

\bibitem[{Meng et~al.(2022)Meng, Bau, Andonian, and Belinkov}]{meng2022locating}
Kevin Meng, David Bau, Alex Andonian, and Yonatan Belinkov. 2022.
\newblock Locating and editing factual associations in gpt.
\newblock \emph{Advances in Neural Information Processing Systems}, 35:17359--17372.

\bibitem[{Meng et~al.(2023)Meng, Sharma, Andonian, Belinkov, and Bau}]{meng2023mass}
Kevin Meng, Arnab~Sen Sharma, Alex~J Andonian, Yonatan Belinkov, and David Bau. 2023.
\newblock Mass-editing memory in a transformer.
\newblock In \emph{The Eleventh International Conference on Learning Representations}.

\bibitem[{Menick et~al.(2022)Menick, Trebacz, Mikulik, Aslanides, Song, Chadwick, Glaese, Young, Campbell-Gillingham, Irving et~al.}]{menick2022teaching}
Jacob Menick, Maja Trebacz, Vladimir Mikulik, John Aslanides, Francis Song, Martin Chadwick, Mia Glaese, Susannah Young, Lucy Campbell-Gillingham, Geoffrey Irving, et~al. 2022.
\newblock Teaching language models to support answers with verified quotes.
\newblock \emph{arXiv preprint arXiv:2203.11147}.

\bibitem[{Mitchell et~al.(2021)Mitchell, Lin, Bosselut, Finn, and Manning}]{mitchell2021fast}
Eric Mitchell, Charles Lin, Antoine Bosselut, Chelsea Finn, and Christopher~D Manning. 2021.
\newblock Fast model editing at scale.
\newblock In \emph{International Conference on Learning Representations}.

\bibitem[{Mitchell et~al.(2022)Mitchell, Lin, Bosselut, Manning, and Finn}]{mitchell2022memory}
Eric Mitchell, Charles Lin, Antoine Bosselut, Christopher~D Manning, and Chelsea Finn. 2022.
\newblock Memory-based model editing at scale.
\newblock In \emph{International Conference on Machine Learning}, pages 15817--15831. PMLR.

\bibitem[{Muhlgay et~al.(2024)Muhlgay, Ram, Magar, Levine, Ratner, Belinkov, Abend, Leyton-Brown, Shashua, and Shoham}]{muhlgay2024generating}
Dor Muhlgay, Ori Ram, Inbal Magar, Yoav Levine, Nir Ratner, Yonatan Belinkov, Omri Abend, Kevin Leyton-Brown, Amnon Shashua, and Yoav Shoham. 2024.
\newblock Generating benchmarks for factuality evaluation of language models.
\newblock In \emph{Proceedings of the 18th Conference of the European Chapter of the Association for Computational Linguistics (Volume 1: Long Papers)}, pages 49--66.

\bibitem[{Ni et~al.(2023)Ni, Chen, Li, Hu, Xu, and Yang}]{ni2023forgetting}
Shiwen Ni, Dingwei Chen, Chengming Li, Xiping Hu, Ruifeng Xu, and Min Yang. 2023.
\newblock Forgetting before learning: Utilizing parametric arithmetic for knowledge updating in large language models.
\newblock \emph{arXiv preprint arXiv:2311.08011}.

\bibitem[{Ouyang et~al.(2022)Ouyang, Wu, Jiang, Almeida, Wainwright, Mishkin, Zhang, Agarwal, Slama, Ray et~al.}]{ouyang2022training}
Long Ouyang, Jeffrey Wu, Xu~Jiang, Diogo Almeida, Carroll Wainwright, Pamela Mishkin, Chong Zhang, Sandhini Agarwal, Katarina Slama, Alex Ray, et~al. 2022.
\newblock Training language models to follow instructions with human feedback.
\newblock \emph{Advances in neural information processing systems}, 35:27730--27744.

\bibitem[{Peng et~al.(2023)Peng, Galley, He, Cheng, Xie, Hu, Huang, Liden, Yu, Chen et~al.}]{peng2023check}
Baolin Peng, Michel Galley, Pengcheng He, Hao Cheng, Yujia Xie, Yu~Hu, Qiuyuan Huang, Lars Liden, Zhou Yu, Weizhu Chen, et~al. 2023.
\newblock Check your facts and try again: Improving large language models with external knowledge and automated feedback.
\newblock \emph{arXiv preprint arXiv:2302.12813}.

\bibitem[{Radford et~al.(2019)Radford, Wu, Child, Luan, Amodei, Sutskever et~al.}]{radford2019language}
Alec Radford, Jeffrey Wu, Rewon Child, David Luan, Dario Amodei, Ilya Sutskever, et~al. 2019.
\newblock Language models are unsupervised multitask learners.
\newblock \emph{OpenAI blog}, 1(8):9.

\bibitem[{Schulman et~al.(2017)Schulman, Wolski, Dhariwal, Radford, and Klimov}]{schulman2017proximal}
John Schulman, Filip Wolski, Prafulla Dhariwal, Alec Radford, and Oleg Klimov. 2017.
\newblock Proximal policy optimization algorithms.
\newblock \emph{arXiv preprint arXiv:1707.06347}.

\bibitem[{Shinn et~al.(2023)Shinn, Labash, and Gopinath}]{shinn2023reflexion}
Noah Shinn, Beck Labash, and Ashwin Gopinath. 2023.
\newblock Reflexion: an autonomous agent with dynamic memory and self-reflection.
\newblock \emph{arXiv preprint arXiv:2303.11366}, 2(5):9.

\bibitem[{Stiennon et~al.(2020)Stiennon, Ouyang, Wu, Ziegler, Lowe, Voss, Radford, Amodei, and Christiano}]{stiennon2020learning}
Nisan Stiennon, Long Ouyang, Jeffrey Wu, Daniel Ziegler, Ryan Lowe, Chelsea Voss, Alec Radford, Dario Amodei, and Paul~F Christiano. 2020.
\newblock Learning to summarize with human feedback.
\newblock \emph{Advances in Neural Information Processing Systems}, 33:3008--3021.

\bibitem[{Touvron et~al.(2023)Touvron, Martin, Stone, Albert, Almahairi, Babaei, Bashlykov, Batra, Bhargava, Bhosale et~al.}]{touvron2023llama}
Hugo Touvron, Louis Martin, Kevin Stone, Peter Albert, Amjad Almahairi, Yasmine Babaei, Nikolay Bashlykov, Soumya Batra, Prajjwal Bhargava, Shruti Bhosale, et~al. 2023.
\newblock Llama 2: Open foundation and fine-tuned chat models.
\newblock \emph{arXiv preprint arXiv:2307.09288}.

\bibitem[{Trivedi et~al.(2023)Trivedi, Balasubramanian, Khot, and Sabharwal}]{trivedi2023interleaving}
Harsh Trivedi, Niranjan Balasubramanian, Tushar Khot, and Ashish Sabharwal. 2023.
\newblock Interleaving retrieval with chain-of-thought reasoning for knowledge-intensive multi-step questions.
\newblock In \emph{Proceedings of the 61st Annual Meeting of the Association for Computational Linguistics (Volume 1: Long Papers)}, pages 10014--10037.

\bibitem[{Wang et~al.(2023)Wang, Yang, Qiu, Liang, He, Gu, Xiao, and Wang}]{wang2023knowledgpt}
Xintao Wang, Qianwen Yang, Yongting Qiu, Jiaqing Liang, Qianyu He, Zhouhong Gu, Yanghua Xiao, and Wei Wang. 2023.
\newblock Knowledgpt: Enhancing large language models with retrieval and storage access on knowledge bases.
\newblock \emph{arXiv preprint arXiv:2308.11761}.

\bibitem[{Ye et~al.(2023)Ye, Liu, Zhang, Hua, and Jia}]{ye2023cognitive}
Hongbin Ye, Tong Liu, Aijia Zhang, Wei Hua, and Weiqiang Jia. 2023.
\newblock Cognitive mirage: A review of hallucinations in large language models.
\newblock \emph{arXiv preprint arXiv:2309.06794}.

\bibitem[{Zhang et~al.(2024)Zhang, Yu, and Feng}]{zhang2024truthx}
Shaolei Zhang, Tian Yu, and Yang Feng. 2024.
\newblock Truthx: Alleviating hallucinations by editing large language models in truthful space.
\newblock \emph{arXiv preprint arXiv:2402.17811}.

\bibitem[{Zhang et~al.(2023{\natexlab{a}})Zhang, Cui, Bi, and Shi}]{zhang2023alleviating}
Yue Zhang, Leyang Cui, Wei Bi, and Shuming Shi. 2023{\natexlab{a}}.
\newblock Alleviating hallucinations of large language models through induced hallucinations.
\newblock \emph{arXiv preprint arXiv:2312.15710}.

\bibitem[{Zhang et~al.(2023{\natexlab{b}})Zhang, Li, Cui, Cai, Liu, Fu, Huang, Zhao, Zhang, Chen et~al.}]{zhang2023siren}
Yue Zhang, Yafu Li, Leyang Cui, Deng Cai, Lemao Liu, Tingchen Fu, Xinting Huang, Enbo Zhao, Yu~Zhang, Yulong Chen, et~al. 2023{\natexlab{b}}.
\newblock Siren's song in the ai ocean: a survey on hallucination in large language models.
\newblock \emph{arXiv preprint arXiv:2309.01219}.

\bibitem[{Zhao et~al.(2023)Zhao, Zhou, Li, Tang, Wang, Hou, Min, Zhang, Zhang, Dong et~al.}]{zhao2023survey}
Wayne~Xin Zhao, Kun Zhou, Junyi Li, Tianyi Tang, Xiaolei Wang, Yupeng Hou, Yingqian Min, Beichen Zhang, Junjie Zhang, Zican Dong, et~al. 2023.
\newblock A survey of large language models.
\newblock \emph{arXiv preprint arXiv:2303.18223}.

\bibitem[{Zheng et~al.(2023)Zheng, Li, Dong, Fan, Wu, Xu, and Chang}]{zheng2023can}
Ce~Zheng, Lei Li, Qingxiu Dong, Yuxuan Fan, Zhiyong Wu, Jingjing Xu, and Baobao Chang. 2023.
\newblock Can we edit factual knowledge by in-context learning?
\newblock In \emph{Proceedings of the 2023 Conference on Empirical Methods in Natural Language Processing}, pages 4862--4876.

\bibitem[{Zheng et~al.(2024)Zheng, Zhang, Zhang, Ye, and Luo}]{zheng2024llamafactory}
Yaowei Zheng, Richong Zhang, Junhao Zhang, Yanhan Ye, and Zheyan Luo. 2024.
\newblock Llamafactory: Unified efficient fine-tuning of 100+ language models.
\newblock \emph{arXiv preprint arXiv:2403.13372}.

\bibitem[{Zhou et~al.(2024)Zhou, Liu, Xu, Iyer, Sun, Mao, Ma, Efrat, Yu, Yu et~al.}]{zhou2024lima}
Chunting Zhou, Pengfei Liu, Puxin Xu, Srinivasan Iyer, Jiao Sun, Yuning Mao, Xuezhe Ma, Avia Efrat, Ping Yu, Lili Yu, et~al. 2024.
\newblock Lima: Less is more for alignment.
\newblock \emph{Advances in Neural Information Processing Systems}, 36.

\end{thebibliography}

\clearpage

\appendix

\section{Appendix}

\subsection{The Algorithm of Proposed Method} \label{111111}

Our method is implemented through Algorithm 1, which is outlined in Section~\ref{Method}.This algorithm provides a step-by-step description of how the LOL framework leverages contrastive decoding between low layers and early exits, alongside the truthfulness refocused module, to alleviate hallucinations in LLMs.

\begin{table}[h] \scriptsize
\setlength\tabcolsep{8pt}
\renewcommand\arraystretch{1.2}
\begin{center}
\begin{tabular}{l}
\toprule[1pt]
Algorithm 1: LOL \\
\hline
\textbf{Input:} original model $\theta$, amateur model $\theta^{*}$, instruction $x_{instruction}$\\ hyperparameters $\lambda$, $\lambda^{'}$, $\lambda^{''}$, $\omega$, $\omega^{'}$, early exiting layer $L$   \\
\textbf{Output:} final logits $\mathcal{F}_{Final}$  \\
1. \textbf{for} token $t$ ← $1$ do    \\
2. \quad logits $\mathcal{F}_t$←$\mathrm{log}p(x_t|x_{<t};\theta)-\lambda\mathrm{log}p(x_t|x_{<t};\theta^{*})$ \\
3. \quad logits $\mathcal{F}_t^{'}$←$\mathrm{log}p(x_t|x_{<t};\theta;L)-\lambda^{'}\mathrm{log}p(x_t|x_{<t};\theta^{*};L)$ \\
4. \quad multi-layer fusion logits:  $\mathcal{F}_{ML}$ ← $\mathcal{F}_t + \omega\mathcal{F}^{'}_t$     \\
5. \quad truthfulness refocused logits: \\  \quad \quad$\mathcal{F}_{TR}$ ← $\mathrm{log}p(x_t|(x_{<t}||x_{instruction});\theta)$  
 \\ \quad \quad \quad \quad \quad \quad $-\lambda^{''}\mathrm{log}p(x_t|(x_{<t}||x_{instruction});\theta^{*})$  \\
6. \quad final logits $\mathcal{F}_{Final}$ ← $\mathcal{F}_{ML} + \omega^{'}\mathcal{F}_{TR}$   \\
7. \quad \textbf{return} $\mathcal{F}_{Final}$  \\

\bottomrule[1pt]

\end{tabular}
\end{center}
\caption{The algorithm for the proposed LOL method.}

\end{table}

\subsection{Completion Details.} \label{hyperparameters}

For our experiments, we leverage LLAMA3-8B-Instruct \cite{llama3modelcard}, Mistral-7B-Instruct-v0.2, and LLAMA2-7B-Chat \cite{touvron2023llama}. To construct the amateur model with hallucinated data, we follow the settings from \cite{zhang2023alleviating} and fine-tune the LLM using the HaluEval dataset.
Specifically, we apply LoRA \cite{hu2021lora} technique for parameter-efficient fine-tuning and hallucination injection. For hardware, we utilize an A800-80G GPU for the experiments. Additionally, we make use of the LLAMA-Factory \cite{zheng2024llamafactory}, a widely adopted and effective framework for fine-tuning, to build our amateur model. The finetuning settings are shown in Table \ref{tab6} for building the amateur model in our experiments.

\begin{table}[h] \small
\setlength\tabcolsep{12pt}
\renewcommand\arraystretch{1.2}
\begin{center}
\begin{tabular}{cc}
\toprule[1pt]
\textbf{Setting} & \textbf{Value} \\
\hline
Model& \makecell[c]{LLAMA2-7B-Base \\LLAMA3-8B-Instruct  \\Mistral-7B-Instruct} \\

Epochs& 5 \\

Device& 4 Nvidia A800 GPU (80GB) \\

Batch size& 256 \\

Learning rate& 5x10$^{-4}$ \\

LoRA Target& $q_{proj}, k_{proj}, v_{proj}$ \\

\bottomrule[1pt]

\end{tabular}
\end{center}
\caption{Finetuning settings for building the amateur model in our experiments across four benchmarks.}
\label{tab6}
\end{table}

\begin{figure}[t]
	\centering
\includegraphics[width=0.9\linewidth]{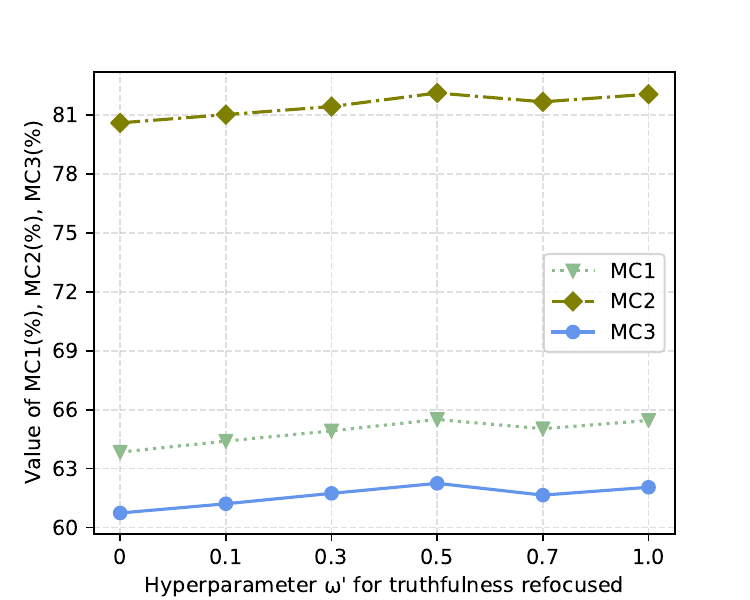}
	\caption{Experimental results of the truthfulness refocused module with different ratios $\omega^{'}$ on TruthfulQA based on LLAMA3-8B-Instruct.}
	\label{f11}
\end{figure}

\subsection{Case Study} \label{case study}

To further demonstrate the effectiveness of our proposed method and explore the knowledge storage of specific early-exit layer, we present a case study examining the generation results of ICD and LOL on StrategyQA, a logic-based generation benchmark. The case details are summarized in Table~\ref{tab4}. Specifically, we show the contrastive decoding results from ICD with early-exit layers (with the \rm{$24th$} layer selected) to show how LOL enhances the final output by leveraging the supplementary knowledge from lower layers. In the first question, both the base model and ICD fail to reason the correct answer. However, contrastive decoding using the $24^{th}$ layer yields the correct result, enabling LOL to produce the correct answer.  Similarly, in the second question, the first three experimental settings fail to generate the correct answer.  Yet, the output from the $24^{th}$ layer contrastive decoding includes some relevant correct  knowledge, which assists LOL in acquiring additional supplementary knowledge, ultimately leading to a correct answer. These cases highlight that LOL improves the factuality of the model through multi-layer fusion and truthfulness capturing during contrastive decoding.

\begin{table*}[t] \footnotesize
\setlength\tabcolsep{6pt}
\renewcommand\arraystretch{1.1}
\begin{center}
\begin{tabular}{p{3.5cm}p{3.5cm}p{3.5cm}p{3.5cm}}
\toprule 
\textbf{Base} & \textbf{Base + ICD} & \textbf{Base + ICD ($24th$ layer)} & \textbf{Base + LOL}\\
\hline
\multicolumn{4}{c}{\textbf{\textit{Question:}} \textit{Are pennies commonly used in Canada?} {\textbf{\textit{Correct Answer:}} \textit{No.}}} \\
\hline
The Canadian penny was discontinued in 2013. Although pennies are no longer in circulation, \textcolor{BrickRed}{they may still be found in circulation}. So the answer is \textcolor{BrickRed}{yes}. & \textcolor{ForestGreen}{Pennies are part of the US currency}. Canada is in North America and has many permanent populations from the US. \textcolor{BrickRed}{Pennies may be used frequently}. So the answer is \textcolor{BrickRed}{yes}. & Canada has its own currency, the Canadian dollar. \textcolor{ForestGreen}{\textbf{Pennies are part of the US currency}}, the US dollar. Thus, \textcolor{ForestGreen}{pennies are not commonly used in Canada}. So the answer is \textcolor{ForestGreen}{no}. & The United States uses pennies, but Canada uses coins called pennies, but \textcolor{ForestGreen}{they are worth 1 cent Canadian, not 1 cent US}. So the answer is \textcolor{ForestGreen}{no}.  \\
\hline
\multicolumn{4}{c}{\textbf{\textit{Question:}} \textit{Is strep throat harmless to singer Rita Ora after her 2020 tonsilitis surgery?} {\textbf{\textit{Correct Answer:}} \textit{No.}}}\\
\hline
Rita Ora had her tonsils removed in 2020. Strep throat is an infection that affects the tonsils. \textcolor{BrickRed}{Since Rita Ora no longer has tonsils, she cannot get strep throat}. So the answer is \textcolor{BrickRed}{yes}. & Rita Ora had her tonsils removed in 2020. \textcolor{ForestGreen}{Trep throat is an infection that affects the tonsils}. \textcolor{BrickRed}{Since Rita Ora no longer has tonsils, she cannot get strep throat}. So the answer is \textcolor{BrickRed}{yes}. So the answer is yes. & Rita Ora had her tonsils removed in 2020. \textcolor{ForestGreen}{\textbf{Tonsils are part of the immune system and help fight infections}}. \textcolor{BrickRed}{Without tonsils, Rita Ora would not be able to get strep throat}. So the answer is \textcolor{BrickRed}{yes}. & Rita Ora had her tonsils removed in 2020. \textcolor{ForestGreen}{Tonsils are part of the immune system and help fight infections. Without tonsils, the immune system is less effective at fighting infections. Strep throat is a bacterial infection. Without tonsils, Rita Ora may be more susceptible to strep throat.} Thus, \textcolor{ForestGreen}{strep throat may not be harmless to Rita Ora}. So the answer is \textcolor{ForestGreen}{no}.  \\

\bottomrule

\end{tabular}
\caption{Case study of LOL, presenting the generation results of hallucination alleviation with ICD and LOL, respectively, based on LLAMA3-8B-Instruct on StrategyQA. \textcolor{ForestGreen}{Green text} denotes factual content, while \textcolor{BrickRed}{red text} indicates hallucinated content.}
\label{tab4}
\end{center}
\end{table*}

\subsection{Theoretical Interpretability of Multi-Layer Fusion}

Being different from Dola, our method treats the lower layers (i.e., early-exit layers) as supplementary knowledge to correct the output logits of the final layer, rather than as objects for contrastive decoding. Specifically, LOL is significantly different from ICD, Dola and other contrastive decoing methods in terms of theory and formula.

Follow the formulas definition in Section \ref{Method}, we obtain the result of multi-layer fusion contrastive decoding as follow steps:
\begin{equation}
\mathcal{F}_t=\mathrm{log}p(x_t|x_{<t};\theta)-\mathrm{log}p(x_t|x_{<t};\theta^{*}),
\end{equation}
\begin{equation}
\mathcal{F}_t^{'}=\mathrm{log} p(x_t|x_{<t};\theta;L)-\mathrm{log}p(x_t|x_{<t};\theta^{*};L),
\end{equation}
\begin{equation}
p(x_t|x_{<t};\theta)=\mathrm{softmax}(\mathrm{logit}_\theta(x_t|x_{<t})),
\end{equation}
\begin{equation}
p(x_t|x_{<t};\theta^{*})=\mathrm{softmax}(\mathrm{logit}_{\theta^{*}}(x_t|x_{<t})).
\end{equation}
\begin{equation}
\mathcal{F}_{ML}=\mathcal{F}_t + \omega\mathcal{F}^{'}_t,
\end{equation}

where ${F}_{ML}$ denotes the result of our multi-layer fusion contrastive decoding, and $\omega \in (0,1]$ is hyperparameter that controls the contribution the lower-layer contrastive decoding result.

Noting that we can transform the formula as follows:

\begin{equation}
\begin{aligned}
\mathcal{F}_{ML}&=\mathcal{F}_t + \omega\mathcal{F}^{'}_t\\
&= (\mathrm{log}p(x_t|x_{<t};\theta)-\mathrm{log}p(x_t|x_{<t};\theta^{*})) \\
+ &(\mathrm{log} p(x_t|x_{<t};\theta;L)-\mathrm{log}p(x_t|x_{<t};\theta^{*};L)) \\
&= (\mathrm{log}p(x_t|x_{<t};\theta)+\mathrm{log} p(x_t|x_{<t};\theta;L))\\-&(\mathrm{log}p(x_t|x_{<t};\theta^{*})
+\mathrm{log}p(x_t|x_{<t};\theta^{*};L))
\end{aligned}
\end{equation}

After we expand the formula and reassemble it, we can find that the formula is transformed into a contrastive decoding form (the operation symbol in the middle is subtraction, not addition used by LOL). This indicates that within the same one LLM (base model or amateur model), \textbf{the logits output by the final layer of the model and that by the early-exit layer are added rather than subtracted, which is fundamentally different from Dola}. Therefore, the essence of LOL's multi-layer fusion is to perform a wider contrastive decoding, using the early-exit layer as supplementary knowledge to correct the final logits.

As illustrated in the above analysis experiment and case study, we empirically the final layer of LLM sometimes contains incorrect facts, while the middle layer may include correct knowledge, but which is lost during transformer processing. In this case, specific early-exit layer is suitable as supplementary knowledge rather than the contrasting distribution. We demonstrate the effectiveness of our proposed LOL in alleviating hallucinations by leveraging early-exit layer as supplementary knowledge through extensive experiments and analysis.


\end{document}